\documentclass[conference]{IEEEtran}
\IEEEoverridecommandlockouts
\usepackage{cite}
\usepackage{amsmath,amssymb,amsfonts}
\usepackage{algorithmic}
\usepackage{graphicx}
\usepackage{textcomp}
\usepackage{xcolor}
\usepackage[colorlinks,
            linkcolor=black,
            anchorcolor=black,
            citecolor=black]{hyperref}
\usepackage{subfigure}
\usepackage{authblk}

\def\BibTeX{{\rm B\kern-.05em{\sc i\kern-.025em b}\kern-.08em
    T\kern-.1667em\lower.7ex\hbox{E}\kern-.125emX}}
\begin{document}

\title{Robust Semantic Segmentation By Dense Fusion Network On Blurred VHR Remote Sensing Images  \\
}

\author[1]{Yi Peng}
\author[3]{Shihao Sun}
\author[1]{Zheng Wang}
\author[2]{Yining Pan}
\author[2]{Ruirui Li
	\thanks{Ruirui Li is the corresponding author (ilydouble@gmail.com).}}
\affil[1]{Shenzhen Institutes of Advanced Technology, CAS, Shenzhen, China\authorcr Email: {\tt e.samuel2.71@gmail.com, zheng.wang@siat.ac.cn}\vspace{1.5ex}}
\affil[2]{College of Information Science \& Technology, Beijing University of Chemical Technology, Beijing, China \authorcr Email: {\tt panyn\_sigrid@163.com,ilydouble@gmail.com} \vspace{1.5ex}} 
\affil[3]{Beijing Futong Dongfang Technology Co, Ltd, Beijing, China \authorcr Email: {\tt cling2sun@foxmail.com} \vspace{-2ex}} 
	
\maketitle

\begin{figure}[htb]
\begin{minipage}[b]{1.0\linewidth}
\centering
\centerline{\includegraphics[width=8cm]{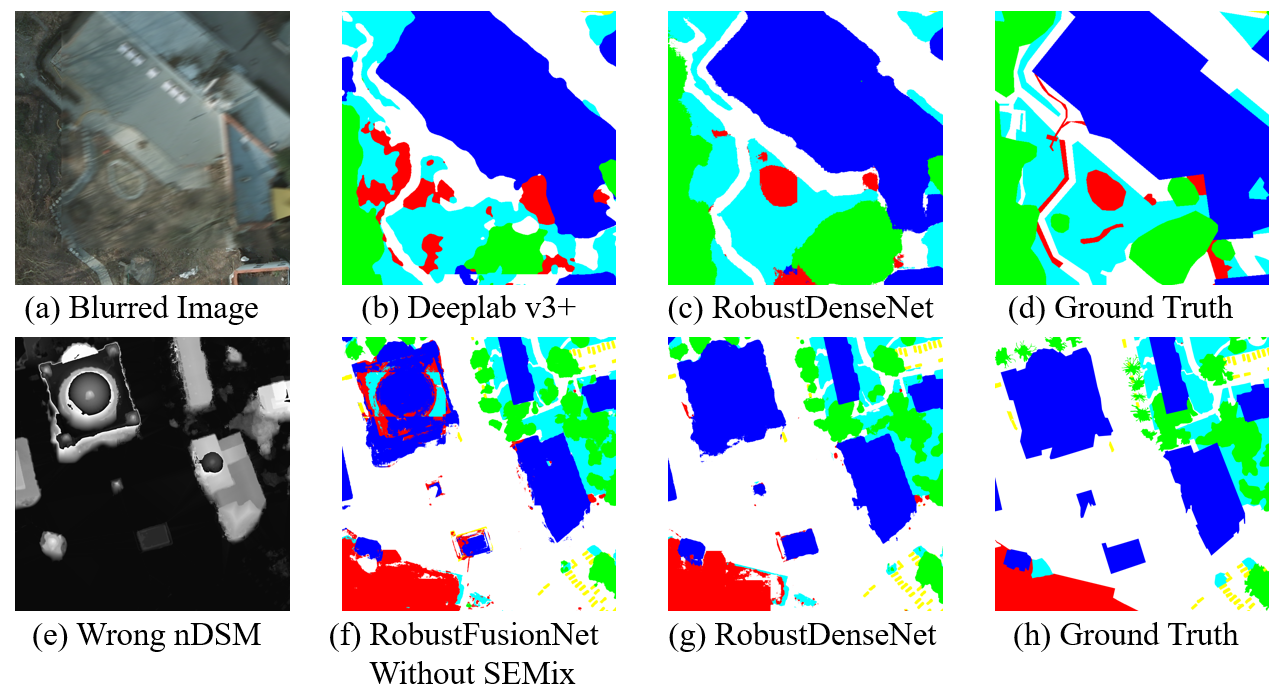}}
\caption{The RobustDenseNet achieves better performance on blurry images compared with Deeplab v3+ and RobustDenseNet without SEMix block.}
\end{minipage}
\label{fig}
\end{figure}

\begin{abstract}
Robust semantic segmentation of VHR remote sensing images from UAV sensors is critical for earth observation, land use, land cover or mapping applications. Several factors such as shadows, weather disruption and camera shakes making this problem highly challenging, especially only using RGB images. In this paper, we propose the use of multi-modality data including NIR, RGB and DSM to increase robustness of segmentation in blurred or partially damaged VHR remote sensing images. By proposing a cascaded dense encoder-decoder network and the SELayer based fusion and assembling techniques, the proposed RobustDenseNet achieves steady performance when the image quality is decreasing, compared with the state-of-the-art semantic segmentation model.
\end{abstract}


\section{Introduction}
Very high resolution(VHR) remotely sensed images, whose ground sampling distance is smaller than 10cm, is an important sort of data for a wide range of applications such as land cover, land use or mapping. These kinds of data can be obtained from unmanned aerial vehicles(UAV) and are often assumed to be with very high image quality in many research work. Semantic segmentation aims to assign pixels in VHR image with category labels, which is an important but unsolved problem in remote sensing. Recently, a lot of deep learning-based methods for VHR remote sensing image segmentation were proposed. For example, modified hourglass networks\cite{DBLP:journals/corr/MarmanisSWGDS16, Volpi2017DenseSL, rs9050446} were applied to handle pixel-level segmentation. To better understand the context, Deeplab v2 and v3\cite{DBLP:journals/corr/abs-1802-02611} take advantage of Atrous Spatial Pyramid Pooling (ASPP) to aggregate features from different resolutions. These methods have already gotten high overall accuracy through supervised learning. However, for images with fuzzy areas and damaged pixels, their performance cannot be guaranteed. Actually, remote sensing images are easily influenced by camera shakes or noise pollution. Information loss can be reconstructed through spatial or spectral clues in multi-modality data. Valada et al.\cite{OB16d} propose the use of the multispectral and multimodal image to increase the robustness of segmentation in the real-world outdoor environment. Remotely sensed images are usually stored in a heterogeneous way, for instance, in the format of NIR, RGB or digital surface model (DSM). Recently, a few researchers\cite{AUDEBERT201820} reported that the classification result could be significantly improved by using the height information (DSM). However, in the aspect of robustness on defective images, fusion of multiple modalities and spectra has not been sufficiently explored in the context of semantic segmentation. 

In this paper, we address the problem of robust segmentation by leveraging dense block and designing a new hourglass convolutional network to obtain pixel-accurate segmentation of defective images. This paper has the following contributions:

\begin{itemize}
\item It designs a new neural network for robust semantic segmentation which leverages cascaded Dense block to aggregate the context information in multimodal channels and uses a novel Up block to eliminate noise during upsampling.
\item It develops the SEMix model to perform a better fusion of DSM and spectral images.
\item A serial of SConv blocks are introduced to learn a dynamic weight for each category, improving the performance on unbalance categories.
\end{itemize}

\section{Methodology}

The main structure of the proposed neural network named RobustDenseNet is shown in Fig 2. Based on the hourglass encoder-decoder structure, the Dense block is used for feature extraction while a modified PixelShuffle structure called Up block is designed for upsampling. Just like \cite{Volpi2017DenseSL}, features output from the Dense block are concatenated to the correspondent Up block to achieve resolution expansion. 

\begin{figure}[htb]
\begin{minipage}[b]{1.0\linewidth}
\centering
\centerline{\includegraphics[width=8.5cm]{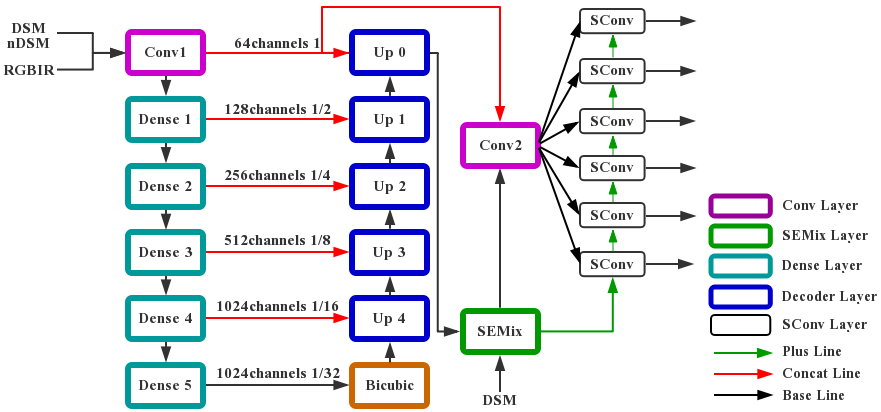}}
\caption{The network structure of the proposed RobustDenseNet.}
\end{minipage}
\end{figure}

The structure of Dense block is the same as that in DenseNet. But their number of channels are different. In this paper, we set the number of channels in our Dense block to 64, 128, 256, 512, 1024, and 1024, respectively, which are proportionally increased by two. The modification helps to increase the accuracy by resampling the features among different semantic levels. 

\begin{figure}[htb]
\begin{minipage}[b]{1.0\linewidth}
\centering
\centerline{\includegraphics[width=8.5cm]{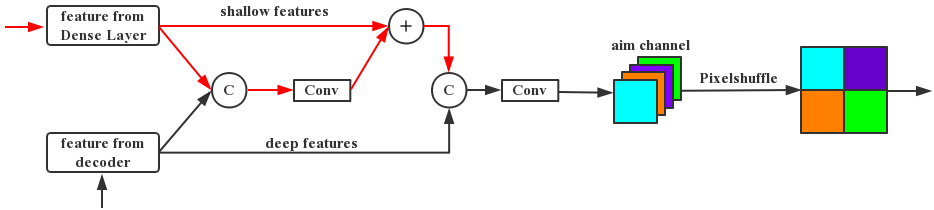}}
\caption{The structure of the Up block in RobustDenseNet. It is added a fusion mechnism to the PixelShffule.}
\end{minipage}
\end{figure}

The Up block is illustrated in Fig.3 in detail. The basic idea of resolution extension by PixelShuffle\cite{DBLP:journals/corr/ShiCHTABRW16} is to directly assemble channels of pixels with low resolution into a two-dimensional block with high resolution. Directly using PixelShuffle without any feature fusion operation leads to ununiform segmentation results as Fig.4(a) shows. It is zoomed out from Fig.4(b).

\begin{figure}[htb]
\begin{minipage}[b]{1.0\linewidth}
\centering
\centerline{\includegraphics[width=8.5cm]{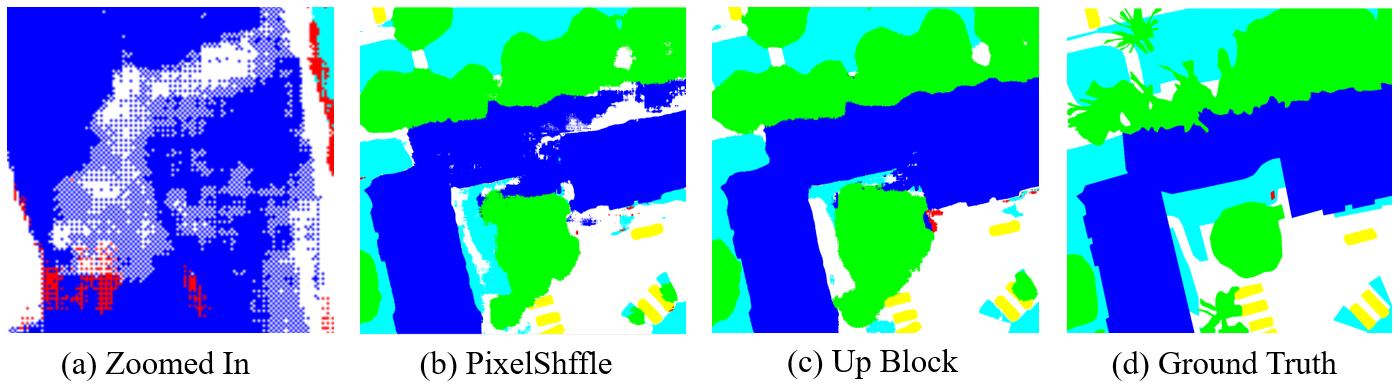}}
\caption{Comparison of results between using Pixelshuffle and Up block in upsampling. }
\end{minipage}
\end{figure}

In order to eliminate the ununiform effects, the Up block first fuses the shallow features with the deep features by a concatenated convolution and a weighted addition. Then the summed features are concatinated to the deep features as the input of the PixelShuffle. Experiments show that the Up block helps to get uniform and continuous segmentation results.

\subsection{SEMix}

The SEMix block is designed for better leveraging the information of DSM images. The DSM images are observed containing imprecise height information(Fig.1e).  This block enhances the robustness by providing a synthetical evaluation on the quality of the data. Its structure is shown in Fig.5.  Inspired by SENet\cite{DBLP:journals/corr/abs-1709-01507}, we use a SELayer which learns proportional weights of DSM.  At the end, the weighted DSM features and the previously learned features are added together to fulfill the fusion.

\begin{figure}[htb]
\begin{minipage}[b]{1.0\linewidth}
\centering
\centerline{\includegraphics[width=8.5cm]{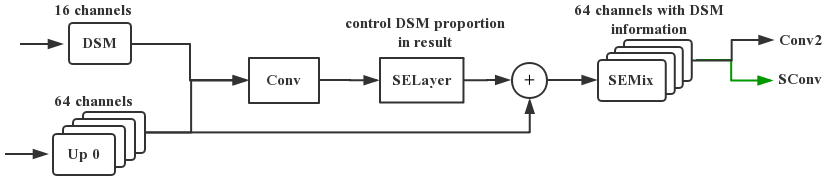}}
\caption{The structure of the SEMix block. }
\end{minipage}
\end{figure}

\subsection{SConv}
The SConv is another critical block that is designed for better balancing categories. As often observed, the semantic information provided by pixels of different categories vary a lot, particularly in blurred or damaged remote sensing images. Take DSM data for example, the imprecise height information play different effects on different categories. Thus each category need a distinct group of weights to evaluate the importance of the features for semantic segmentation. In the SConv, we also use the SELayer to learn the weights of the features.

Every SConv is mapped to a category. Assuming $i$ is the class No., the output of the SConv is represented by $x$ and $w$ is the learnt weights, the pseudo code is illustrated in $Algorithm$ 1.


\begin{figure}[htb]
\begin{minipage}[b]{1.0\linewidth}
\centering
\centerline{\includegraphics[width=8.5cm]{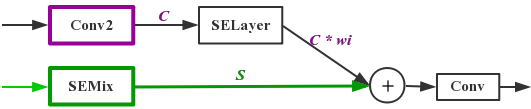}}
\caption{The structure of the SConv block.}
\end{minipage}
\end{figure}

\begin{tabular}{l}
\hline
\textbf{Algorithm 1}  SConv\\
\hline
1:\textbf{For {each $i\in [1, classNumber]$} }\\
2:\quad\quad $C*w_i$ = SELayer(C)\\
3:\quad\quad $x_i$ = Conv(C * $w_i$ + S)\\
4: \textbf{end for}\\
\hline
\label{algo:scheduling}
\end{tabular}

We use Cross Entropy as loss function for RobustDenseNet. It uses output $x$ counted by $Algorithm$ 1 and true label $i$ to calculate loss value. It is defined in (1).

\begin{equation}
\begin{aligned}
loss(x, i)=-x_{i} + log\biggl(\sum_{j}e^{x_j}\biggr)
\end{aligned}
\end{equation}

\section{Experiments And Analysis}
To validate the proposed DenseFusionNet, we do experiments on the ISPRS potsdam 2D dataset and compare the DenseFusionNet with the state-of-the-art deep model Deeplab v3+ in the same environments. The Potsdam dataset contains 38 orthorectified tiles (of the same size) of size 6000×6000 pixels with a spatial resolution of 5 cm, over the town of Potsdam (Germany). We choose 18 tiles as training set, 5 tiles as validation set and 14 tiles for testing. We randomly add motion blur to spectra data as well as randomly delete colors of small areas to simulate the real cases in the wild. In the experiment, we try to contaminate pixels in proportion to area size, from none to 50\%. The training tiles are augmented by rotation and are clipped into 1280*1280 patches during training and testing.


All the experiments are carried out on a laboratory computer. Its GPU is NVIDIA GTX 1080Ti which have 10GB memory. In order to use GPU more efficiently for training large network, such as Deeplab v3+, we use group normalization\cite{DBLP:journals/corr/abs-1803-08494} with single batch instead of batch normalization. The main required packages include python 3.6, CUDA8.0, cudnn7, pyTorch0.4.1 and etc. 

The performance is evaluated on the blurred or partial damaged dataset for overall accuracy (OA). To evaluate class-specific performance, the F1 metric is used, computed as the harmonic mean between precision and recall.

\begin{figure}[htb]
\begin{minipage}[b]{1.0\linewidth}
\centering
\centerline{\includegraphics[width=8.5cm]{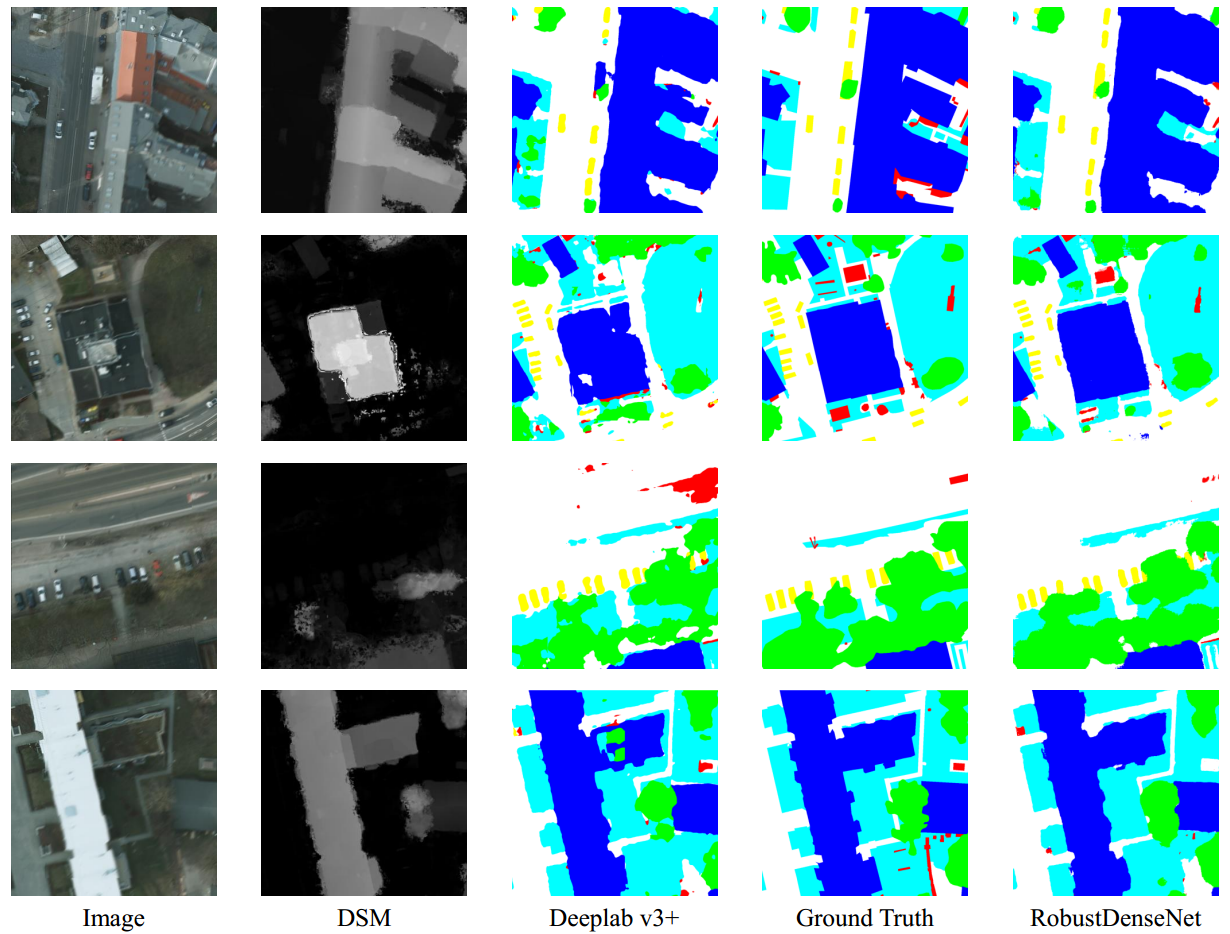}}
\caption{Quantitative comparisons between Deeplab v3+ and the RobustDenseNet with different damaged conditions.}
\end{minipage}
\end{figure}

The experimental results are listed in Table1. As the area of the blurred or damaged pixels increases (from none to 50\%), the proposed RobustDenseNet achieved steady performances on both overall accuracy and F1 metrics. The overall accuracy decreased slightly from 90.3 to 89.8, which is probably caused by the decline on Car class (from 94.7 to 94.0) and the Imp-suf class (from 92.7 to 91.8). On the contrary, Deeplab v3+, even as the latest model in semantic segmentation, failed in robustness testing. Its mean F1 score collapsed from 88.6 to 86.2 and the overall accuracy decreased by 2.4\%.  Fig.8 shows the changing trend on overall accuracy and F1 metric, as the area of blurred or damaged pixels increases. It is observed that the RobustDenseNet decreases more slowly than the Deeplab V3+.

In the fuzzy images, Imp\_suf pixels are easily misclassified to the clutter. By fusing DSM data through the RobustDenseNet, segmetation results of our method have less error than that of Deeplab v3+. This phenominone can be obeserved in Fig.7 (red areas). The building pixels and car pixels in blurred images tent to loss sharp edges. The results on 50\% blurry images are showed in Fig.1 and Fig.7. The segmentation results usually contain bubble effects which are reliefed by RobustDenseNet through SEMix and SConv. These structures provide a selective fusion mechnism.

\begin{table}[htbp]
\caption{The quantitative comparison between Deeplab v3+ and the RobustDenseNet in different degree of image damage. }
\begin{center}
\begin{tabular}{|c|c|c|c|c|c|c|}
\hline
\textbf{Method} & \multicolumn{3}{|c|}{\textbf{Deeplab v3+}} & \multicolumn{3}{|c|}{\textbf{RobustDenseNet}} \\
\hline
Degree & {\textbf{0\%}} & {\textbf{20\%}} & {\textbf{50\%}} & {\textbf{0\%}} & {\textbf{20\%}} & {\textbf{50\%}} \\
\hline
Imp-suf & 91.0 & 90.5 & 89.2 & 92.7 & 92.4 & 91.8\\
\hline
Building & 96.2 & 96.2 & 96.2 & 97.2 & 96.8 & 97.1\\
\hline
Low-veg & 83.3 & 82.3 & 79.9 & 86.7 & 86.4 & 86.1\\
\hline
Tree & 81.6 & 81.2 & 76.6 & 86.7 & 85.7 & 86.4\\
\hline
Car & 91.0 & 90.8  & 89.0 & 94.7 & 94.1 & 94.0\\
\hline
OA & 87.6 & 87.1 & 85.4 & 90.3 & 89.9 & 89.8\\
\hline
Mean F1 & 88.6  & 88.2 & 86.2 & 91.6 & 91.0 & 91.0\\
\hline
\end{tabular}
\label{table:benchmarks}
\end{center}
\end{table}

\begin{figure}[htb]
\begin{minipage}[b]{1.0\linewidth}
\centering
\centerline{\includegraphics[width=8.5cm]{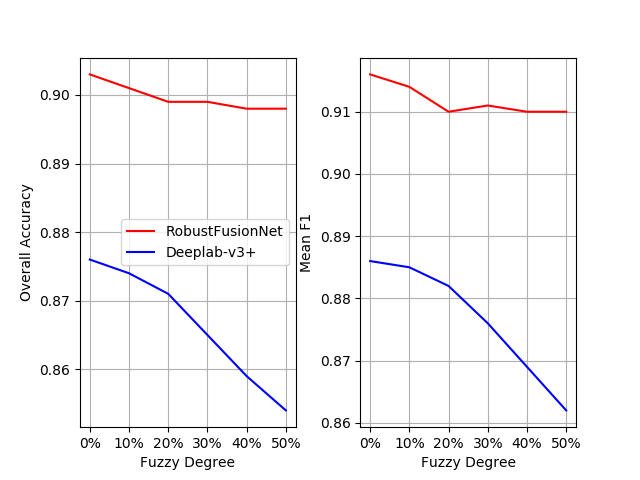}}
\caption{OA and Mean F1 curves of different fuzzy degree between deeplab v3+ and RobustDenseNet on ISPRS 2D labeling potsdam test set.}
\end{minipage}
\end{figure}

\section{conclusion}
In summary, this paper proposed a dense block based  hourglass network for robust semantic segmentation. The robustness is come from three improvements of the network structure: 1) cascaded dense block for feature extraction;  2) SEMix for better fusing DSM data; 3) SConv for learning category-aware weights of features. In the experiments on the blurred and patially damaged VHR remotely sensed image, the proposed RobustDenseNet achieves steady performance on blurred images. Compared with the state-of-the-art model Deeplab v3+, the decrease of overall accuracy and F1 metrics are much slower.

\bibliographystyle{unsrt}
\bibliography{references}

\end{document}